\documentclass[11pt]{article}
\usepackage{acl2016}
\usepackage{times}
\usepackage{latexsym}
\usepackage{url}
\usepackage{algorithm}
\usepackage{algpseudocode}
\usepackage{pifont}
\usepackage{times}
\usepackage{url}
\usepackage{multirow}
\usepackage{amssymb}
\usepackage{amsthm}
\usepackage{amsmath}
\usepackage{verbatim}
\usepackage{tikz-qtree}
\usepackage{tikz}

\algrenewcomment[1]{\(\triangleright\) #1}
\algnewcommand{\LineComment}[1]{\State \(\triangleright\) #1}

\newcommand{\ignore}[1]{}

\usepackage[T1]{fontenc}

\aclfinalcopy % Uncomment this line for the final submission
\def\aclpaperid{xxx} %  Enter the acl Paper ID here

\title{Problems With Evaluation of Word Embeddings\\
Using Word Similarity Tasks}
\author{Manaal Faruqui$^1$ ~ Yulia Tsvetkov$^1$ ~ Pushpendre Rastogi$^2$ ~  Chris Dyer$^1$\\
$^1$Language Technologies Institute, Carnegie Mellon University\\
$^2$Department of Computer Science, Johns Hopkins University\\
{\tt \{mfaruqui,ytsvetko,cdyer\}@cs.cmu.edu, pushpendre@jhu.edu}
}

\begin{document}

\maketitle

\begin{abstract}
Lacking standardized extrinsic evaluation methods for vector
representations of words, the NLP community has relied heavily on \textit{word
similarity} tasks as a proxy for intrinsic evaluation of word vectors.
Word similarity evaluation, which correlates the distance between vectors and
human judgments of ``semantic similarity'' is attractive, because it is
computationally inexpensive and fast.  In this paper we present several
problems associated with the evaluation of word vectors on word similarity
datasets, and summarize existing solutions. Our study suggests that the use of
word similarity tasks for evaluation of word vectors is not sustainable and
calls for further research on evaluation methods.
\end{abstract}

\section{Introduction}

Despite the ubiquity of word vector representations in NLP,
there is no consensus in the community on
what is the best way for evaluating word vectors. The most popular intrinsic
evaluation task is the \textit{word similarity} evaluation. In word similarity
evaluation, a list of pairs of words along with their similarity rating
(as judged by human annotators) is provided. The task is to measure how well
the notion of word similarity according to humans is captured by the word
vector representations. Table~\ref{tab:example} shows some word pairs along
with their similarity judgments from WS-353 \cite{finkelstein:2002}, a popular
word similarity dataset.

Let $a$, $b$ be two words, and $\mathbf{a}, \mathbf{b} \in \mathbb{R}^{D}$ be
their corresponding word vectors in a $D$-dimensional vector space.
Word similarity in the vector-space can be obtained by computing the
cosine similarity between the word vectors of a pair of words:
\begin{equation}
\label{eq:cos}
\textrm{cosine}(\mathbf{a}, \mathbf{b}) = \frac{\mathbf{a} \cdot
\mathbf{b}}{\Vert \mathbf{a} \Vert \, \Vert \mathbf{b} \Vert}
\end{equation}
where, $\Vert \mathbf{a} \Vert$ is the $\ell_2$-norm of the vector, and
$\mathbf{a} \cdot \mathbf{b}$ is the dot product of the two vectors.
Once the vector-space similarity between the words is computed, we obtain
the lists of pairs of words sorted according to vector-space similarity, and
human similarity. Computing Spearman's correlation \cite{citeulike:8703893}
between these ranked lists provides some insight into how well the learned word vectors capture intuitive notions of word
similarity.
\begin{table}[!tb]
  \centering
  \small
  \begin{tabular}{l|l|c}
    %\hline
Word$_1$ & Word$_2$ & Similarity score [0,10]\\
\hline
love & sex & 6.77\\
stock & jaguar & 0.92\\
money & cash & 9.15\\
development & issue & 3.97\\
lad & brother & 4.46\\
%\hline
    \end{tabular}
  \caption{Sample word pairs along with their human similarity judgment from
WS-353.}
  \label{tab:example}
\end{table}

Word similarity evaluation is attractive, because it is computationally
inexpensive and fast, leading to faster prototyping and development of word
vector models. The origin of word similarity tasks can be tracked back to
\newcite{Rubenstein:1965} who constructed a list of 65 word pairs with
annotations of human similarity judgment. They created this dataset to
validate the veracity of the distributional hypothesis \cite{harris-54}
according to which the meaning of words is evidenced by the context they
occur in. They found a positive correlation between contextual similarity and
human-annotated similarity of word pairs. %\mfar{motivation behind WS353}
%\yt{the last sentence of the first paragraph goes here}
Since then, the lack of a standard evaluation method for word vectors has led to
the creation of several \textit{ad hoc} word similarity datasets.
Table~\ref{tab:wordsimbench} provides a list of such benchmarks obtained from
\url{wordvectors.org} \cite{faruqui:2014:demo}.

In this paper, we give a comprehensive analysis of the problems that are
associated with the evaluation of word vector representations using word
similarity tasks.\footnote{
An alternative to correlation-based word similarity evaluation is the \textit{word analogy} task,
where the task is to find the missing word $b^{*}$ in the relation: $a$ is to
$a^{*}$ as $b$ is to $b^{*}$, where $a$, $a^{*}$ are related by the same relation as $a$, $a^{*}$. For example, $king:man :: queen:woman$.
\newcite{mikolov-yih-zweig:2013:NAACL} showed that this problem can be solved using the vector offset method:
$\mathbf{b}^{*} \approx \mathbf{b} - \mathbf{a} + \mathbf{a}^{*}$.
\newcite{levy2014linguistic} show that
solving this equation is equivalent to computing a linear combination of word
similarities between the query word $b^{*}$, with the given words $a$, $b$, and $b^{*}$. Thus, the results we present in this paper naturally extend to the word analogy tasks.}
We survey existing literature to construct a list of such
problems and also summarize existing solutions to some of the problems.
%suggest solutions to address them whenever possible.
Our findings suggest that word similarity tasks are not appropriate for
evaluating word vector representations, and call for further research on
better evaluation methods

\begin{table}[!tb]
  \centering
  \small
  \begin{tabular}{lrl}
    %\hline
Dataset & Word pairs & Reference\\
\hline
RG & 65 & \tiny \newcite{Rubenstein:1965}\\
MC & 30 & \newcite{mc:1991} \\
WS-353 & 353 & \newcite{finkelstein:2002}\\
YP-130 & 130 & \newcite{Yang06verbsimilarity}\\
MTurk-287 & 287 & \newcite{Radinsky:2011:WTC:1963405.1963455}\\
MTurk-771 & 771 & \newcite{halawi2012large}\\
MEN & 3000 & \newcite{bruni:2012}\\
RW & 2034 & \newcite{Luong-etal:conll13:morpho}\\
Verb & 144 & \newcite{baker2014unsupervised}\\
SimLex & 999 & \newcite{HillRK14}\\
%\hline
    \end{tabular}
  \caption{Word similarity datasets.}
  \label{tab:wordsimbench}
\end{table}

\section{Problems}

We now discuss the major issues with evaluation of word vectors using word
similarity tasks, and present existing solutions (if available) to address them.

\subsection{Subjectivity of the task}

The notion of word similarity is subjective and is often confused with
relatedness. For example, \textit{cup}, and \textit{coffee} are related to
each other, but not similar.
\textit{Coffee} refers to a plant (a living organism)
or a hot brown drink, whereas \textit{cup} is a man-made object,
%\yt{which may contain coffee.},
which contains liquids, often coffee.
Nevertheless, \textit{cup} and \textit{coffee} are rated more similar than
pairs such as \textit{car} and \textit{train} in WS-353
\cite{finkelstein:2002}. Such anomalies are also found in recently constructed
datasets like MEN \cite{bruni:2012}. Thus, such datasets unfairly penalize
word vector models that capture the fact that \textit{cup} and \textit{coffee}
are dissimilar.

In an attempt to address this limitation, \newcite{agirre:2009} divided WS-353 into two sets containing word pairs
exhibiting only either similarity or relatedness. Recently, \newcite{HillRK14}
constructed a new word similarity dataset (SimLex), which captures the degree
of similarity between words, and related words are considered dissimilar.
Even though it is useful to separate the concept of similarity and relatedness,
it is not clear as to which one should the word vector models be expected to
capture. %\yt{the critique is valid, then you present hill's solution to it. but the last sentence is subjective and not very clear, maybe remove it? }

\subsection{Semantic or task-specific similarity?}
Distributional word vector models capture some aspect of word
co-occurrence statistics of the words in a language
\cite{levy2014neural,levy2015improving}. Therefore, to the extent these models
produce semantically coherent representations, it
%\yt{rephrase "this" -- it's not clear what you are referring to}
can be seen as evidence of the distributional hypothesis of
\newcite{harris-54}. Thus, word embeddings like Skip-gram, CBOW, Glove, LSA
\cite{Turney10fromfrequency,mikolov:2013,glove:2014} which are trained on word
co-occurrence counts can be expected to capture semantic word similarity, and
hence can be evaluated on word similarity tasks.

Word vector representations which are trained as part of a neural network
to solve a particular task (apart from word co-occurrence prediction) are called
distributed word embeddings \cite{collobert:2008}, and they are task-specific
in nature. These embeddings capture task-specific word similarity, for example,
if the task is of POS tagging, two nouns \textit{cat} and \textit{man}
might be considered similar by the model, even though they are not semantically
similar. Thus, evaluating such task-specific word embeddings on word similarity
can unfairly penalize them. This raises the question: what kind of word
similarity should be captured by the model?

\subsection{No standardized splits \& overfitting}

To obtain generalizable machine learning models, it is necessary to make sure
that they do not overfit to a given dataset. Thus, the datasets are usually
partitioned into a training, development and test set on which the model is
trained, tuned and finally evaluated, respectively \cite{Manning:1999}.
Existing word similarity datasets are not partitioned into training,
development and test sets. Therefore, optimizing the word vectors to
perform better at a word similarity task implicitly \textit{tunes on the test
set} and overfits the vectors to the task. %, which is undesirable.
On the other hand, if researchers decide to perform their own splits of the
data, the results obtained across different studies can be incomparable.
Furthermore, the average number of word pairs in the word
similarity datasets is small ($\approx 781$, cf. Table~\ref{tab:wordsimbench}),
and partitioning them further into smaller subsets may produce unstable
results.  %with unstable estimates.

We now present some of the solutions suggested by previous work to avoid
overfitting of word vectors to word similarity tasks.
\newcite{faruqui-dyer:2014:EACL2014}, and \newcite{Lu2015DeepMC} evaluate the
word embeddings exclusively on word similarity and word analogy tasks.
\newcite{faruqui-dyer:2014:EACL2014} tune their embedding on one word
similarity task and evaluate them on all other tasks. This
%\yt{precludes test/train pollution.}
ensures that their vectors are being evaluated on held-out datasets.
\newcite{Lu2015DeepMC} propose to directly evaluate the generalization of a
model by measuring the performance of a single model on a large gamut of tasks.
This evaluation can be performed in two different ways: (1) choose the
hyperparameters with best average performance across all tasks, (2) choose
the hyperparameters that beat the baseline vectors on most
tasks.\footnote{Baseline vectors can be any off-the-shelf vector models.}
By selecting the hyperparameters that perform well across a range of tasks,
these methods ensure that the obtained vectors are generalizable.
\newcite{stratos2015model} divided each word similarity dataset
individually into tuning and test set and reported results on the test set.
% \yt{remove "across different tasks" (repetitive) }across different tasks.

\subsection{Low correlation with extrinsic evaluation}
Word similarity evaluation measures how well the notion of word similarity
according to humans is captured in the vector-space word representations.
Word vectors that can capture word similarity might be expected to perform
well on tasks that require a notion of explicit semantic similarity between
words like paraphrasing, entailment. However, it has been shown that no strong
correlation is found between the performance of word vectors on word similarity
and extrinsic evaluation NLP tasks like text classification, parsing, sentiment
analysis
\cite{tsvetkov:2015:eval,schnabel-2015}.\footnote{In these studies, extrinsic
evaluation tasks are those tasks that use the dimensions of word vectors as
features in a machine learning model. The model learns weights for how important
these features are for the extrinsic task.}
An absence of strong correlation between the word similarity evaluation and
downstream tasks calls for alternative approaches to evaluation.

\subsection{Absence of statistical significance}
There has been a consistent omission of statistical significance for measuring
the difference in performance of two vector models on word similarity tasks.
Statistical significance testing is important for validating metric gains
in NLP \cite{berg2012empirical,sogaard2014s}, specifically while solving
non-convex objectives where results obtained due to optimizer instability can
often lead to incorrect inferences \cite{clark2011better}.
%\yt{remove 1 sentence; remove paragraph break}
The problem of statistical significance in word similarity evaluation
was first systematically addressed by \newcite{walid:15}, who used Steiger's
test \cite{steiger1980}\footnote{A quick tutorial on Steiger's test \& scripts:
\url{http://www.philippsinger.info/?p=347}} to compute how significant the
difference between rankings produced by two different models is against the
gold ranking. However, their method needs explicit ranked list of words
produced by the models and cannot work when provided only with the correlation
ratio of each model with the gold ranking. This problem was solved by
\newcite{rastogi2015multiview}, which we describe next.

\newcite{rastogi2015multiview}
observed that the improvements shown on small word similarity task datasets by
previous work were insignificant. We now briefly describe the method presented
by them to compute statistical significance for word similarity evaluation.
Let $A$ and $B$ be the rankings produced by two word vector models over a list
of words pairs, and $T$ be the human annotated ranking. Let $r_{AT}$, $r_{BT}$
and $r_{AB}$ denote the Spearman's correlation between $A:T$, $B:T$ and $A:B$
resp. and $\hat{r}_{AT}$, $\hat{r}_{BT}$ and $\hat{r}_{AB}$ be their empirical
estimates. \newcite{rastogi2015multiview} introduce $\sigma_{p_0}^{r}$ as the
minimum required difference for significance (MRDS) which satisfies the
following:
\begin{equation}
(r_{AB} < r) \land (|\hat{r}_{BT} - \hat{r}_{AT}| < \sigma_{p_0}^{r})
\implies pval > p_0
\end{equation}
Here $pval$ is the probability of the test statistic under the null hypothesis
that $r_{AT} = r_{BT}$ found using the Steiger's test.
The above conditional ensures that
if the empirical difference between the
rank correlations of the scores of the competing methods to the gold ratings is
less than $\sigma_{p_0}^{r}$
then either the true correlation between the competing methods is greater than $r$,
or the null hypothesis of no difference has p-value greater than $p_0$.
$\sigma_{p_0}^{r}$ depends on the size of the dataset, $p_0$ and $r$ and
\newcite{rastogi2015multiview} present its values for common word similarity
datasets.
%\footnote{Please refer to the original paper by \newcite{steiger1980} on how to compute these statistics.}
Reporting statistical significance in this way would help estimate
the differences between word vector models.

\subsection{Frequency effects in cosine similarity}
The most common method of measuring the similarity between two words in the
vector-space is to compute the cosine similarity between the corresponding word
vectors.
Cosine similarity implicitly measures the similarity between two unit-length
vectors (eq.~\ref{eq:cos}). This prevents any biases in favor of frequent words
which are longer as they are updated more often during training
\cite{turian:2010}.

Ideally, if the geometry of embedding space is primarily driven by semantics,
the relatively small number of frequent words should be evenly distributed
through the space, while large number of rare words should cluster around
related, but more frequent words.
However, it has been shown that vector-spaces contain \textit{hubs}, which are
vectors that are close to a large number of other vectors in the space
\cite{radovanovic2010hubs}. This problem manifests in word vector-spaces
in the form of words that have high cosine similarity with a
large number of other words \cite{dinu2014improving}.
\newcite{schnabel-2015} further refine this \textit{hubness} problem to show that
there exists a power-law relationship between the frequency-rank\footnote{The
rank of a word in vocabulary of the corpus sorted in decreasing order of
frequency.} of a word and the frequency-rank of its neighbors.
Specifically, they showed that the average rank of the 1000 nearest neighbors
of a word follows:
\begin{equation}
\textrm{nn-rank} \approx 1000 \cdot \textrm{word-rank}^{0.17}
\end{equation}

This shows that pairs of words which have similar frequency will be
closer in the vector-space, thus showing higher word similarity than they
should according to their word meaning. Even though newer datasets of word
similarity sample words from different frequency
bins \cite{Luong-etal:conll13:morpho,HillRK14}, this still does not
solve the problem that cosine similarity in the vector-space gets polluted by
frequency-based effects. Different distance normalization schemes have been
proposed to downplay the frequency/hubness effect when computing
nearest neighbors in the vector space
\cite{dinu2014improving,tomavsev2011probabilistic}, but their applicability as
an absolute measure of distance for word similarity tasks still needs to
investigated.

\subsection{Inability to account for polysemy}
Many words have more than one meaning in a language.
For example, the word \textit{bank} can either correspond to a financial
institution or to the land near a river. However in WS-353, \textit{bank}
is given a similarity score of $8.5/10$ to \textit{money}, signifying that
\textit{bank} is a financial institution.
%its considered to have the sense of a financial institution.
Such an assumption
of one sense per word is prevalent in many of the existing word similarity
tasks, and it can incorrectly penalize a word vector model for capturing
a specific sense of the word absent in the word similarity task.

To account for sense-specific word similarity, \newcite{huang2012improving}
introduced the Stanford contextual word similarity dataset (SCWS), in which the
task is to compute similarity between two words given the contexts they occur
in. For example, the words \textit{bank} and \textit{money} should have a low
similarity score given the contexts: \textit{``along the east \underline{bank}
of the river''}, and \textit{``the basis of all \underline{money} laundering''}. Using cues from
the word's context, the correct word-sense can be identified and the appropriate
word vector can be used.
Unfortunately, word senses are also ignored by majority of the frequently used
word vector
models like Skip-gram and Glove. However, there has been progress on obtaining
multiple vectors per word-type to account for different word-senses
\cite{reisinger:naaclx10,huang2012improving,neelakantan-EtAl:2014:EMNLP2014,jauhar:2015,rothe}.

\section{Conclusion}

In this paper we have identified problems associated with word similarity
evaluation of word vector models, and reviewed existing solutions wherever possible.
Our study suggests that the use of word similarity tasks for evaluation of word
vectors can lead to incorrect inferences and calls for further research on
evaluation methods.

Until a better solution is found for intrinsic evaluation of
word vectors, we suggest task-specific evaluation: word vector models should be
compared on how well they can perform on a downstream NLP task. Although
task-specific evaluation produces different rankings of word vector models for
different tasks \cite{schnabel-2015}, this is not necessarily a problem
because different vector models capture different types of information which can
be more or less useful for a particular task.

\bibliography{references}
\bibliographystyle{acl2016}
\end{document}